\documentclass[sn-mathphys]{sn-jnl}

\theoremstyle{thmstyleone}%
\theoremstyle{thmstyletwo}%

\theoremstyle{thmstylethree}%

\newcommand{\rev}[1]{\textcolor{black}{#1}}
\newcommand{\qilong}[1]{\textcolor{blue}{#1}}

\begin{document}

\title[On Mask-based Image Set Desensitization...]{On Mask-based Image Set Desensitization with Recognition Support}


\author[1]{\fnm{Qilong} \sur{Li}}
\equalcont{These authors contributed equally to this work.}

\author*[2]{\fnm{Ji} \sur{Liu}}\email{jiliuwork@gmail.com}
\equalcont{These authors contributed equally to this work.}

\author[3]{\fnm{Yifan} \sur{Sun}}

\author*[1]{\fnm{Chongsheng} \sur{Zhang}}\email{cszhang@henu.edu.cn}

\author[4]{\fnm{Dejing} \sur{Dou}}

\affil[1]{\orgdiv{School of Computer and information Engineering}, \orgname{Henan University}, \orgaddress{\city{Kaifeng}, \postcode{475004}, \country{China}}}

\affil[2]{\orgname{Hithink RoyalFlush Information Network Co., Ltd.}, \orgaddress{\city{Hangzhou}, \postcode{310023}, \country{China}}}

\affil[3]{\orgname{Baidu Inc.}, \orgaddress{\city{Beijing}, \postcode{100193}, \country{China}}}

\affil[4]{\orgname{Boston Consulting Group}, \orgaddress{\city{Beijing}, \postcode{100020}, \country{China}}}



\abstract{In recent years, Deep Neural Networks (DNN) have emerged as a practical method for image recognition. The raw data, which contain sensitive information, are generally exploited within the training process. However, when the training process is outsourced to a third-party organization, the raw data should be desensitized before being transferred to protect sensitive information. Although masks are widely applied to hide important sensitive information, preventing inpainting masked images is critical, which may restore the sensitive information. The corresponding models should be adjusted for the masked images to reduce the degradation of the performance for recognition or classification tasks due to the desensitization of images. In this paper, we propose a mask-based image desensitization approach while supporting recognition. This approach consists of a mask generation algorithm and a model adjustment method. We propose exploiting an interpretation algorithm to maintain critical information for the recognition task in the mask generation algorithm. In addition, we propose a feature selection masknet as the model adjustment method to improve the performance based on the masked images. Extensive experimentation results based on multiple image datasets reveal significant advantages (up to 9.34\% in terms of accuracy) of our approach for image desensitization while supporting recognition.}

\keywords{Pattern recognition, Object recognition, Deep learning, Image mask, Recognition support}



\maketitle

\section{Introduction}
\label{sec:intro}
\rev{
The explosive development of social media and the growing usage of camera and video devices has generated a huge number of images, e.g., face images, plant images, animal images, etc. The raw data is  generally directly leveraged to train Deep Neural Networks (DNN), e.g., Facenet \cite{schroff2015facenet}, CosFace \cite{wang2018cosface}, and ArcFace \cite{deng2019arcface}, without considering the privacy of sensitive information within the images while incurring potential legal risks. The raw data may contain critical sensitive information, which requires careful protection. In addition, the raw data may have authority or ethical issues, such as MS1M \cite{2016MS}, discouraging the sharing or releasing of the datasets \cite{chen2021perceptual, murgia2019microsoft}. Furthermore, because of cost control or technique limit, an organization may need to outsource the model design and training process to a third-party organization, while protecting the sensitive information. End users may submit an image dataset to Machine Learning as a Service (MLaaS) to get a trained model for a specific task without revealing critical information. Thus, it is critical to train models while protecting the sensitive information within the raw image dataset without degrading the performance of the trained models \cite{raynal2020image}.
}

\begin{figure}[t]
  \centering
  \includegraphics[scale=.75]{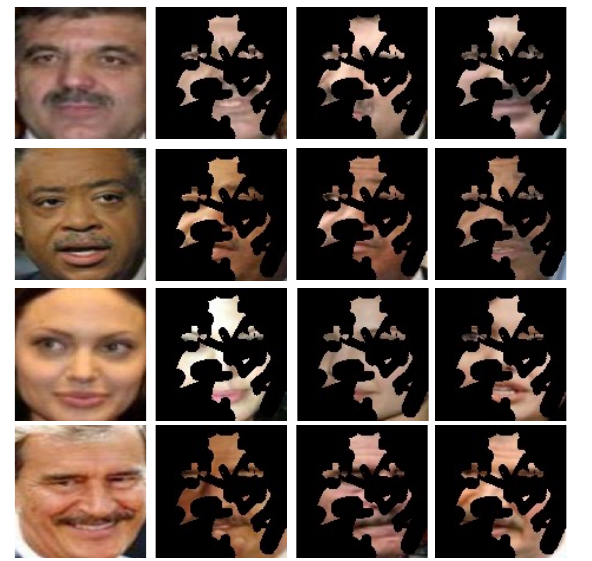}
  \caption{\textbf{Mask-based image desensitization.} First column contains the raw face image. The right three columns contain the masked images. The images of the same row correspond to the same person.}
  \label{fig:illustration_method}
\end{figure}

In order to protect the privacy of sensitive information within images, it is beneficial to desensitize the raw data to train a DNN model with the minimal useful information for a specific task. The desensitization of images refers to removing sensitive information while ensuring sufficient remaining information for Machine Learning (ML) tasks. The primary objective of image desensitization is to destroy the visual characteristics of the raw images while preventing inpainting or restoring the original images from the desensitized data \cite{castellanos2010data,bakken2004data}. In addition, the remaining information within the desensitized images should not negatively impact the accuracy of the trained ML models, the structure of which can be adjusted to the desensitized images.

\rev{While the devices occupy large amount of storage, a large amount of personal data is stored on the powerful devices, and users are often unwilling to directly provide raw images to model training. In order to leverage these vast amounts of distributed data on devices, federated learning emerges as a promising approach to enable the collaborative training without transferring the raw data. \cite{mcmahan2021advances}. While a bunch of techniques \cite{liu2022Efficient,Zhang2022FedDUAP,Li2022FedHiSyn} have been proposed, federated learning directly exploit the use of raw data or with simple differential privacy protection. In this paper, we incorporate a data desensitization process on the local terminal with the federated learning, to further ensure the privacy of sensitive personal data.}  

Although many methods have been proposed to protect the privacy of images, the existing methods seldom consider the impact of the accuracy of models while desensitizing the raw images. Some encryption methods, e.g., homomorphic encryption \cite{aono2016scalable, gentry2009fully}, can well protect the privacy, it suffers either a complicated adaptation of the current ML models to the encrypted data or high cost to train a model based on encrypted data \cite{Nandakumar2019Towards}. While they can provide privacy notions, Differential Privacy (DP) methods \cite{dwork2014algorithmic} or image obfuscation methods \cite{chen2021perceptual} can hardly destroy the visual characteristics of images, which may either degrade the accuracy of ML models or make it easy to recover sensitive information. \rev{Blurring, mosaic, and binarization masks are three commonly used image desensitization techniques. By applying these techniques to the raw image, the sensitive information of the image is significantly distorted, thereby achieving visual desensitization. However, since blurring and mosaic techniques operate globally on images, they not only destroy sensitive information but also distort non-sensitive information, which is detrimental to training ML models \cite{cciftcci2017reliable}. Compared to the aforementioned techniques, binarization mask has the advantage of operating only on specific regions of an image. Therefore, if properly designed to mask only certain specific areas, it can effectively reduce the impact on training ML models while preserving necessary useful information \cite{wilber2016can}. In order to effectively utilize desensitized data for training ML models, some algorithms have been proposed. Face recognition with binarization masks \cite{elmahmudi2019deep} is a pioneering approach to understanding the impact of masked images on training ML models, while this method only utilizes some simple and fixed mask shapes with low recognition accuracy and a large area of occlusion in masked images.} In addition, if these desensitization methods are not well designed, the desensitized images may be inpainted \cite{li2020recurrent} or restored \cite{xing2021end, zhang2020deblurring}. 

\rev{In this paper, we aim to explore the design of an image desensitization approach that can effectively mask sensitive information without compromising ML model training. While there are existing binary mask generation algorithms \cite{liu2018partialinpainting, yu2019free}, they are primarily designed for inpainting techniques and lack targeted desensitization operations on images. We propose a novel image desensitization approach to enable the training process with masked images to prevent the easy restoration of the desensitized images while guaranteeing the performance of a specific designed network.}

In this paper, we propose an efficient mask-based image set desensitization approach while ensuring the performance of ML models. \rev{The approach consists of a mask generation method and Feature Select Masknet (FSM). The mask generation method generates a mask template with minimal preserved visual information from the original image, ensuring both irreversibility and data availability with the assistance of an interpretable model \cite{zhang2018top}. Specifically, a series of candidate occlusions are generated through the help of the interpretable model, and then the most suitable occlusion image is selected based on our proposed selection criteria. FSM is an embeddable module that can be integrated into any network structure with the objective to filter the features generated from the masked regions so as to prevent incorrect features from affecting the training process of the model.} An example of mask-based image desensitization is illustrated in {Fig.~\ref{fig:illustration_method}}. We summarize our main contributions as follows:
\begin{enumerate}
    \item \rev{We propose a mask-based image desensitization method, which generates a series of candidate mask images with the assistance of an interpretable model. Then, using our proposed selection algorithm, we obtain the most suitable mask image.}
    \item \rev{We propose a Feature Selection Masknet (FSM), which can be integrated into any backbone network to handle masked images. This network is designed to be used in conjunction with the mask template. By incorporating this network into an ML model, it enables the filtering of features generated from the masked regions, thereby enhancing the usability of desensitized data during the training process.}
    \item We conduct extensive experiments based on multiple datasets to validate our desensitization approach and show the advantages of our approach compared with existing approaches.
\end{enumerate}
The rest of this paper is organized as follows. In {Section~\ref{sec:related}}, we present literature in relevant areas. {Section~\ref{sec:method}} details our proposed approach, including the mask generation method, an adjusted model with masked images, and the exploitation of federated learning for distributed data. In {Section~\ref{sec:experiments}}, we present extensive experimentation to validate our approach. {Section~\ref{sec:conclusion}} gives the conclusion.

\section{Related Work}
\label{sec:related}

Our approach is mainly related to image desensitization, privacy protection, interpretation, inpainting, recognition with occlusions, and federated learning. 

\textbf{Image desensitization.} Three widely used desensitization approaches include mask, blurring, and pixelation. Mask refers to replacing certain regions, masked with the same shape as the original image, with a predefined color, e.g., black \cite{elmahmudi2019deep, liu2018partialinpainting}. Blurring switches the value of each pixel by a Gaussian average of the neighbors, which generally reduces the high-frequency information of the raw images \cite{flusser2015recognition}. Pixelation divides a region into multiple non-overlapping rectangular grids while the color of each grid is set to the average color or original pixels \cite{fan2018image}. Although these approaches can protect the privacy of images, a few methods have been proposed to restore the original images. For instance, super-resolution techniques are proposed to recover the images with mosaic (pixelation) \cite{xing2021end}, deblurring methods exist for the blurred images \cite{zhang2020deblurring}, and inpainting techniques can be used to restore the masked images \cite{li2020recurrent}. Compared with the blurring and pixelation, the masking method does not rely on a sliding window and can remarkably destroy the image while preventing inpainting or restoration \qilong{when mask areas are big enough} \cite{wilber2016can}. 
\qilong{The target of the proposed mask-based desensitization method is to destroy the maximum areas in the origin image while maintaining the data availability. Compared to the existing mask-generated methods \cite{elmahmudi2019deep, liu2018partialinpainting}, our proposed mask-based desensitization method generates masks in a searching manner with the guide of interpretation method, which obtains more mask shape than \cite{elmahmudi2019deep} and more mask areas than \cite{elmahmudi2019deep, liu2018partialinpainting} corresponding to higher dSSIM and accuracy while protecting the privacy of the original images.}

\textbf{Privacy protection.} Encryption methods, e.g., homomorphic encryption \cite{aono2016scalable, gentry2009fully, paillier1999public,lei2018practical}, can encrypt the raw data to protect privacy. However, the training process should be adapted to the encrypted data, which may incur expensive overhead \cite{Nandakumar2019Towards}. In addition, the adaptation is much more complicated as well. Differential Privacy (DP) can add noises to the raw data to protect privacy \cite{dwork2014algorithmic,abadi2016deep,yan2022monitoring,li2022protecting}. However, it cannot remove visual characteristics of the original images, which may have limited protection on privacy or degrade the accuracy of ML models.

Some methods \cite{li2021identity, li2021learning, wen2022identitymask, cui2021multitask, yuan2022pro, ijcai2018p91, mirjalili2020privacynet} desensitize face images based on the ideas of face manipulation. In these methods, some face attributes are modified so that the changed face turns into ``another'' person, but they retain the availability of manipulated images. Our  method has the same purpose as these methods, i.e., desensitizing face images on the visual level but still maintaining data availability. However, compared with these methods, our desensitization method is practical and can destroy the visual information while maintaining excellent recognition capability.

\textbf{Interpretation.} Interpretation tools can be utilized to reveal the importance of training data for the inference of ML models \cite{koh2017understanding, li2021interpretable}. The regions with a significant contribution to the performance of ML models can be calculated using the top-down attention mechanism, while this approach cannot foreground the significant impact of relevant regions. Excitation BackProp (EBP) \cite{zhang2018top} adopts contrastive top-down attention to differentiate the relevant areas where the object is located from the outside areas, the attention of which can well represent the contributions. 

\textbf{Image Inpainting.} Inpainting techniques can be exploited to recover defects of limited sizes. Deep Learning (DL)-based inpainting techniques have achieved significant advances \cite{liu2018image, liu2019coherent, jo2019sc, yu2019free}. The progressive inpainting method is a widely used type of DL-based inpainting technique, which consists of two sequential steps, i.e., edge restoration and color restoration \cite{nazeri2019edgeconnect, xiong2019foreground, li2019progressive}. However, these two steps are within different feature spaces, and the color restoration may be restricted by the edge restoration, which incurs inconsistency. As proposed in RFR-Net \cite{li2020recurrent}, the inpainting process can also be divided into multiple steps, each of which deals with a specific area of the masked region. With RFR-Net, the inpainting process of one area restores the edge and the color simultaneously, corresponding to better consistency results than the progressive method. In addition, the inpainted results of one area can be used for the inpainting process of other areas to improve consistency further.

\textbf{Recognition with occlusions.} Besides the inpainting techniques, some approaches have been proposed to realize the recognition based on images with occlusions. For instance, recognition performance can be improved by increasing the dimension of hidden representation and selecting essential neurons in the head layers in DeepID2+ \cite{sun2015deeply}. However, it is hard to combine DeepID2+ with other models. While attributing high weights to the non-occluded parts and low weights to the occluded parts can deal with the occlusions, combined with existing models \cite{wan2017occlusion}, it cannot identify the influence of added noises brought by the convolutions layers within the original model. While some occluded areas are useless or harmful to the performance of CNN models, the corrupted feature elements are eliminated to improve the accuracy \cite{song2019occlusion}, which does not exploit the mask information. 
\qilong{Inspired by these methods, in this paper, we design a feature selection masknet, which aims to eliminate the features generated in the mask areas in the recognition model. Different from \cite{wan2017occlusion, song2019occlusion}, the proposed feature selection masknet is customized to enhance the recognition accuracy of the masked images and depends on the mask generated by the proposed mask-based desensitization method. In addition, the proposed feature selection mechanism limits the gated value to only occurring in the mask areas to prevent corrupting the feature in the known areas, which corresponds to higher accuracy of the trained model.}

\textbf{Federal learning.} Federated learning \cite{mcmahan2017communication,liu2022distributed,Liu2022From,mcmahan2021advances} aims to collaboratively train a global model with distributed data while protecting data privacy. Only the gradients or models are allowed to be transferred, and the raw data is kept in local devices. Existing approaches \cite{liu2022Efficient,Zhang2022FedDUAP,Li2022FedHiSyn,Li2020,liu2022distributed,liu2022multi,che2022federated,che2023fast,che2023federated,liu2023distributed,liu2023heterps} mainly focus on improving the accuracy in diverse environments. In this paper, we combine federated learning and image desensitization to protect data privacy further.

\section{Methodology}
\label{sec:method}
In this section, we illustrate our proposed method for mask-based image desensitization with recognition support. First, we formulate the problem to address. Then, we propose the image desensitization method. Afterward, we detail the Feature Selection Masknet (FSM) to deal with the desensitized images. In addition, we combine the image desensitization approach with federated learning to handle distributed data. 

\subsection{Problem formulation.}
\label{subsec:problem_define}

We formulate the mask-based image desensitization problem as Formula \ref{eq:problem}. 
\begin{gather}
{arg}\mathop{min} \limits_{\mathcal{D}} \underset{(x,y)\sim\mathcal{(X, Y)}}{\mathbb{E}} ( \mathcal{F}(\mathcal{M}(\mathcal{D}(x)), y) - \gamma \cdot \mathcal{P}(\mathcal{D}(x)) )
\label{eq:problem}
\end{gather}
where $\mathcal{(X, Y)}$ refers to the original image dataset $\mathcal{X}$ and corresponding label set $\mathcal{Y}$, \rev{$\mathcal{M}$ is a given DNN model,} $\mathcal{F}(\cdot,\cdot)$ represents the loss function of the DNN model $\mathcal{M}$, $\mathcal{D}$ represents the desensitization function, $y$ is the label corresponding to a raw image $x$, $\mathcal{P}$ refers to the capacity to recover or inpaint the desensitized image, and $\gamma$ is the importance of preventing inpainting the desensitized images. In this paper, we exploit masks to hide sensitive information. A mask $m$ is a matrix composed of binary values, e.g., 0 or 1, with the same shape as the original image matrix $x$. Then, we can get the desensitized image $\widetilde x$ by the Hadamard product of $m$ and $x$: $\widetilde x = x \odot m$. As a bigger ratio of the masks corresponds to a smaller capacity to inpaint masked images \cite{li2020recurrent}, we utilize the ratio between the size of masked areas and the size of the image to quantify the capacity to recover or inpaint desensitized images, i.e., $\mathcal{P}(\cdot)$ defined in Formula \ref{eq:ratio}.
\begin{gather}
\mathcal{P}(\mathcal{D}(x)) = \frac{||m||_1}{h * w}, 
\label{eq:ratio}
\end{gather}
where $||m||_1$ represents the L1-norm of the mask matrix $m$, and image $x$ is of dimension $h * w$. 

\subsection{Mask-based desensitization}
\label{subsec:mask_search}


The ideal desensitization function $ \mathcal{D} $ is to directly generate a mask $m$ corresponding to an input face image $x$. However, due to the sparsity and lack of ground truth in the binary matrix mask $m$, it is challenging to directly train a deep neural network to obtain the desensitization function $ \mathcal{D}$. Thus, we transform the original objective, which aims to obtain a desensitization function $ \mathcal{D}$ in Formula \ref{eq:problem}, to the objective of obtaining the optimal mask $\widetilde m$ as defined in Formula \ref{eq:mask_optimize}. Then, we can progressively generate candidate masks and search for an optimal candidate mask.
\begin{gather}
\label{eq:mask_optimize}
{\widetilde m} = \mathop{argmin} \limits_{m} \underset{(x,y)\sim\mathcal{(X, Y)}}{\mathbb{E}} (\mathcal{F}(\mathcal{M}(\widetilde x), y) - \gamma \cdot \mathcal{P}(m)), ~~m\sim{m_{can}}
\end{gather}
\rev{Where $m_{can}=\{m_{0},m_{1},...m_{n}\}$ is the candidate set of masks, and the number of candidates is $n$. It is worth noting that the optimization objective given by Formula \ref{eq:mask_optimize} can theoretically be applied to any given DNN model and loss function. In this paper, we adopt the commonly used ResNet \cite{he2016deep} as the given network $\mathcal M$ and employ cross-entropy loss as the loss function to optimize the objective defined in Formula \ref{eq:mask_optimize}.} Through the first item of Formula \ref{eq:mask_optimize}, the searched mask will not impede recognition accuracy, and the second item of Formula \ref{eq:mask_optimize} will destroy the most image area. 

\rev{While existing methods \cite{liu2018partialinpainting, yu2019free} support random generation of masks, directly optimizing Formula \ref{eq:mask_optimize} based on these methods can be computationally expensive and time-consuming due to the uncertainty in generating occluded regions.} To reduce the time consumption of the search procedure and improve search efficiency, the critical area of face images that significantly contribute to the face recognition model should be reserved. In the aligned face dataset, different images have similar critical areas. Thus we can exploit one template for all face images. Therefore, before the mask search procedure, we utilize the explainable face recognition (XFR) method EBP \cite{zhang2018top} to obtain the critical area. In order to generate the template, $K$ face images are sampled from the original aligned image dataset. Then, we calculate the saliency maps, which can highlight the important regions for recognition, with the EBP method $e$. Afterward, the \textbf{Mean Saliency Map (MSM)}, which is obtained by multiple image sampling, is calculated as a template. Furthermore, we choose the regions with importance within the MSM greater than a predefined threshold $T$. \rev{If we increase the value of T, the preserved important regions will decrease; conversely, if we decrease the value of T, the preserved important regions will increase.}. Finally, we generate the critical template $S$ with each element defined in Formula~\ref{eq:xfr_mean}.
\begin{align}
\mathrm{s_{i,j}} = 
\begin{cases}
1, ~~~~~~\textrm{~if~} \frac{1}{K} \sum_{k=1}^{K} {e}(x^k_{i,j}) > T \\
0, ~~~~~~otherwise.
\end{cases}
\label{eq:xfr_mean}
\end{align}
$K$ represents the number of all sampled images with the index $k$, $i$ and $j$ represent the region's index, $1$ represents unmasked regions, and $0$ refers to the masked regions.

\begin{algorithm}[!t]
  \caption{Mask-based desensitization algorithm.}
  \label{algos:mask_search}
  \begin{algorithmic}
    \State \textbf{Input:}
    \State $x, y$: The input face image and label.
    \State $n$: The hyper-parameter controls the number of searched masks.
    \State $K$: The number of salience maps.
    \State $T$: The threshold of binarization.
    \State $v_0, v_1,l,b,a$: The hyper-parameters control to mask generated.
    \State \textbf{Output:}
    \State $\widetilde m$: The searched mask.
    \State $S \gets MSM(K,x)$ according to Formula \ref{eq:xfr_mean}
    \For{$i$ in $\{1, 2, ..., n\}$}
        \State $m_{r} \gets RMG(v_0, v_1,l,b,a)$
        \State $m_{ca} \gets m_{r} + S$
        \State $\widetilde x \gets x \odot m_{ca}$
        \State $O_{ca} \gets \mathcal{F}(\mathcal{M}(\widetilde x), y) - \gamma \cdot \mathcal{P}(m_{ca}) $
        \If{$ O_{ca} < O $}
            \State Updated the optimal mask $\widetilde m$
            \State $O \gets O_{ca}$
            \State $\widetilde m \gets m_{ca}$
        \EndIf
        \State Randomly generate ${v_0, v_1,l,b,a}$
    \EndFor
  \end{algorithmic}
\end{algorithm}

The mask generation process consists of two stages. At the random mask generation stage (Stage I), we exploit a state-of-the-art \textbf{Random Mask Generation (RMG)} algorithm \cite{yu2019free} with 5 hyper-parameters, i.e., min vertex, max vertex, max length, max brush width, and max angle. By controlling these hyper-parameters, we can control the range of the ratio of the randomly generated masks ($m_r$). Then, the candidate is updated by adding the critical template $S$ to $m_r$ to avoid destroying the critical area in face images, as shown in Formula \ref{eq:mask_xfr}.
\begin{equation}
m_{ca} = m_{r} + S
\label{eq:mask_xfr}
\end{equation}

At the select mask stage (Stage II), we need to train a face recognition model for each candidate while minimizing the $\mathcal{F}(\widetilde x, y)$. Then, we generate the proper mask ($\widetilde m$), which can minimize Formula \ref{eq:mask_optimize}. 

It is worth noting that, as enumerators candidates are generated at Stage I, the time consumption of the multiple face recognition model training can be unacceptable. However, we observe a positive correlation between the model trained on raw images and that on desensitized images with the same mask (as shown in Tabel~\ref{tab:mask_level_test}). According to Tabel~\ref{tab:mask_level_test}, on the clean model (the model training on raw images), the test accuracy increases gradually with the increase of mask ratio; on the mask model (the model training on masked images), the test accuracy also increases with the increase of mask ratio. Therefore, there is a positive correlation between the clean and mask models on the loss with masked images. Thus, we utilize the model pre-trained on the clean images as the model used in $ \mathcal{F} $ to reduce the time consumption. The mask-based desensitization algorithm is shown in {Algorithm~\ref{algos:mask_search}}.


\begin{figure*}[t]
  \centering
  \includegraphics[width=0.8\linewidth]{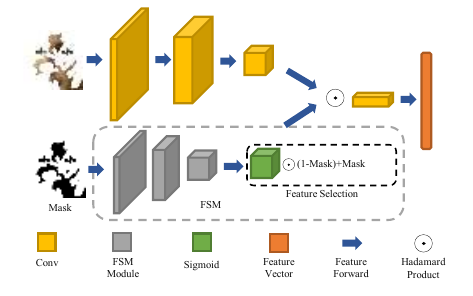}
  \caption{Face recognition with FSM.}
  \label{fig:fsm_net}
\end{figure*}

\subsection{Face Recognition with Pixel Masking}
\label{subsec:fsm}

After generating a proper mask with the minimum impact on image recognition, we designed a model to improve the recognition accuracy with desensitized images. The forward propagation of the recognition model can be seen as a feature extractor. The face recognition network extracts practical features when the input is an uncorrupted image. When the input image is covered by a mask, in the forward propagation, missing regions will be restored by convolution layer by layer. However, the pixels filled in missing areas may harm the recognition. Therefore, we propose a Feature Select Masknet (FSM) to generate an adapted weight matrix with the same shape as the feature map of a backbone network. 

The proposed FSM is to stick several FSM modules into a backbone network. A basic FSM module consists of a convolution layer with a stride of 2, a batch normalization layer, and a relu activation layer. As shown in {Fig.~\ref{fig:fsm_net}}, the input of FSM is the generated proper mask. Within the output of FSM, each pixel is normalized, representing the weight value of the feature within the backbone network. 

Although the feature within the backbone network can be enhanced by multiplying the output of FSM, the unmasked areas in the feature will also be changed by multiplying the weight value. Therefore, feature selection (FS) is necessary to prevent the change of the unmasked areas in the backbone’s feature. FS is formulated as: 
\begin{equation}
\label{eq:fsm}
FS = {fsm}(m) \odot (1-m_{resized})+m_{resized}
\end{equation}
where $m_{resized}$ represents the resized mask according to the feature's shape within the backbone network. The first item in Formula \ref{eq:fsm} can retain the weight value in masked areas while abandoning these in unmasked areas. Then, the weight value in unmasked areas is set to 1 by adding the resized mask. The feature selection mechanism can automatically select efficient features from FSM by end-to-end learning.

\subsection{Train the method with federated learning}
\label{subsec:fsmFL}
Images containing sensitive information incur severe privacy issues to directly transfer raw data among diverse data owners. To alleviate the privacy problem and to obtain an excellent masked recognition model with distributed, we use the above methods with federated learning. The details of federated learning are shown in {Section~\ref{sec:experiments}}.

\section{Experiments}

\label{sec:experiments}
\begin{figure}[t]
  \centering
  \includegraphics[width=.6\linewidth]{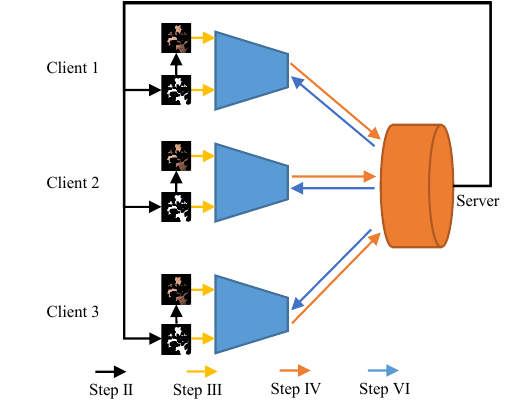}
  \caption{Face recognition with pixel masking based federated learning.}
  \label{fig:fl_masked_recognition}
\end{figure}

In this section, we present the experimental evaluation to show the advantages of our proposed methods. We take a baseline model, which exploits an existing mask dataset (Nvidia mask dataset) according to the Formula \ref{eq:mask_optimize}. In addition, we take existing desensitization methods, i.e., mask, blurring, and mosaic, as baselines. Then, we compare Algorithm \ref{algos:mask_search} with the baseline methods.  

\subsection{Datasets and Experimental Setups}
\label{subsec:exp_settings}
\textbf{Datasets:}~ We conduct experiments based on three datasets as follows:
\begin{itemize}
    \item Vggface2 \cite{cao2018vggface2}. This dataset is collected through Google Image Search and contains 300 million images of more than 9,000 people with large variations in pose, age, and background. We use all images as the training dataset in all models mentioned below.
    \item Lfw \cite{LFWTech}. The lfw (labeled faces in the wild) dataset is often utilized as a test dataset for the face recognition algorithm. The dataset collected more than 13,000 face images from the web, each of which is labeled with a name. In this dataset, 1,680 people have more than two different images. The publisher provides a test pair that can be used to compare face recognition algorithms.
    \item NVIDIA mask dataset \cite{liu2018partialinpainting}. Nvidia is a binarized mask dataset, which contains 12000 random masks with shapes (256, 256, 3). These images are divided into 6 levels according to hole-to-image ratios.
\end{itemize}

\textbf{Face recognition model:}~ We exploit a resnet (residual network) \cite{he2016deep}with 34 layers as the backbone framework and cross-entropy loss to train a face recognition model. According to the training data and the architecture of the model, we denote these models as resnet, resnet-mask, and resnet-fsm as follows:
\begin{itemize}
	\item Resnet represent the model trained based on raw images in vggface2 as training data and resnet-34 as the backbone.
	\item Resnet-mask refers to the model based on masked images as training data and resnet-34 as model.
	\item Resnet-fsm is trained with masked images as training data and exploiting FSM (see details in section~\ref{subsec:fsm}) in the backbone (resnet-34). 
\end{itemize}

\textbf{Implementation details:} The three models are trained with 1 million iterations; the optimizer is Stochastic Gradient Descent (SGD); momentum is set to 0.9, and the learning rate is started with 0.01 and reduced by 10$\times$ at 50\%, 70\%, and 90\% of totally iterations, respectively. For FSM, the size of the input mask matrix is the same as the input image's size (224 $\times$ 224) in the backbone net. The resized shape of the masks ($m_{resized}$)  depends on the FSM insertion point in the backbone net. The resized size is 56, 28, 14, and 7 according to the insertion point after layer1, layer2, layer3, and layer4, respectively. In the training process, we add a classification head after the final avgpool output of the resnet. In the inference process, the classification head is dropped out, and the final output of the avgpool layer is extracted as a feature. The cosine distance between two features is utilized to decide whether they belong to the same person.

\textbf{Federated learning settings:} In our experiment, we train all models with a federated learning setting. We divide the training dataset into three parts on average, which are stored on three different machines. Each machine (client) has 4 NVIDIA Tesla V100 32G, and the training procedure is conducted using the federated learning method of transferring gradients between client and server. The procedure of federated learning is as follows: 

Step I, federated learning initialization. The server exploits Algorithm \ref{algos:mask_search} to obtain the proper mask on a large-scale open-face dataset. Then, the images are aligned with each client. Afterward, global and local models are initialized according to the model proposed in {Section~\ref{subsec:fsm}}. Step II: Each client downloads the mask from the server to desensitize the local data. Step III: The local model in each client is trained by masked face images and the mask download from the server. Step IV: Upload model parameters from each client to the server. Step V: The server integrates all model parameters and updates the global model. Step VI: Each client downloads the latest model parameters from the server to update the local model. Repeat steps 2 to 6 until convergence.

\begin{figure}[t]
  \centering
  \includegraphics[width=0.8\linewidth]{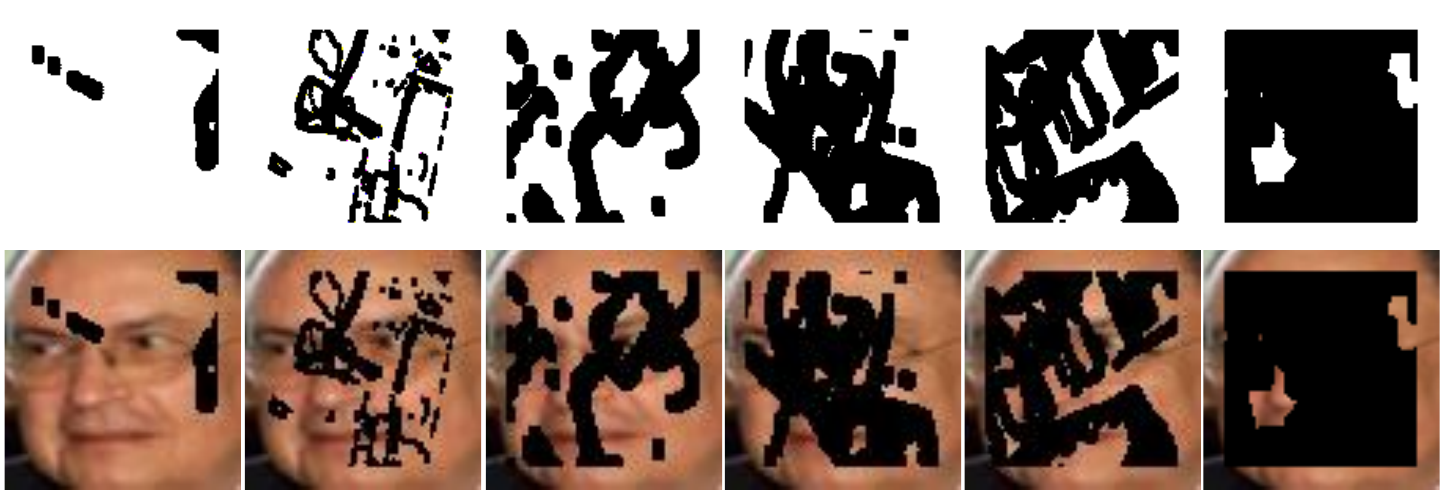}
  \caption{Masks sampled from NVIDIA mask dataset and corresponding masked face. In first row, the image from left to right are the mask sampled from level (0.01-0.1), (0.1-0.2), (0.2-0.3), (0.3-0.4), (0.4-0.5), (0.5-0.6). The second row shows the masked face images. [Best viewed with zoom-in.]}
  \label{fig:nvidia_mask}
\end{figure}

\begin{table}[t]
  \centering
  \caption{The face recognition accuracy of resnet and resnet-mask with different mask levels.}
  \begin{tabular}{@{}lcc@{}c@{}}
    \toprule
    Mask level & resnet & resnet-mask & $\delta$ \\
    \midrule
    None & 93.2\% & 92.13\% & -1.07\% \\
    0.01-0.1 & 90.3\% & 91.89\% & 1.59\%\\
    0.1-0.2 & 83.71\% & 91.32\% & 7.61\%\\
    0.2-0.3 & 73.54\% & 90.49\% & 16.95\%\\
    0.3-0.4 & 62.08\% & 89.53\% & 27.45\%\\
    0.4-0.5 & 56.03\% & 87.76\% & 31.73\%\\
    0.5-0.6 & 53.63\% & 83.62\% & 29.99\%\\
    \bottomrule
  \end{tabular}
  \label{tab:mask_level_test}
\end{table}

\subsection{Baseline Method}
\label{subsec:baseline}
To explore the influence of recognition accuracy with a masked face, we design a baseline model which directly utilizes Formula \ref{eq:mask_optimize} to search for qualified masks on NVIDIA mask dataset. As introduced in {Section~\ref{subsec:exp_settings}}, resnet is directly trained based on raw vggface2 dataset with the cross-entropy loss. With resnet-mask, we randomly sample some masks from NVIDIA mask dataset to the training data in each batch and exploit these masked face images to train the backbone network with the same optimizer and parameters as resnet. Some masked trained data are shown in {Fig.~\ref{fig:nvidia_mask}}.

In the case of masked image recognition, we random sample masks in different mask levels at the test procedure. Then, these masks are applied in raw test images from lfw datasets, and the masked face images are sent to resnet and resnet-mask models. We repeat this experiment 5 times and report the average recognition accuracy. The result is shown in {Table~\ref{tab:mask_level_test}}, where $\delta$ represents the difference of the recognition accuracy of resnet and resnet-mask. 

As shown in {Table~\ref{tab:mask_level_test}}, adding masks to the training data can improve the accuracy. For instance, the recognition accuracy for the mask level (0.5$-$0.6) is 53.63\% with resnet while it becomes 83.62\% with resnet-mask. This indicates that the backbone network can recognize masked images when masked images are utilized within the training procedure. 

\begin{figure}[t]
  \centering
  \includegraphics[width=0.4\linewidth]{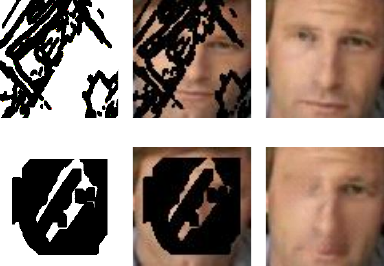}
  \caption{Two masks searched according to Formula \ref{eq:mask_optimize} in mask level (0.4-0.5) and (0.5-0.6). The left column shows the masks, the middle column shows the face with the corresponding mask, and the right column shows the inpainting results.}
  \label{fig:new_gated_conv_mask}
\end{figure}

In order to search for the mask that image recognition models can recognize with maximum mask ratio in NVIDIA mask dataset, we exploit Formula \ref{eq:mask_optimize} to search for a proper mask at mask level (0.5-0.6). We also conduct the same experiment in mask level (0.4-0.5) to verify if a better mask exists. After testing 2000 masks from level (0.4-0.5) and 2000 masks from level (0.5-0.6), we get two proper masks as shown in {Fig.~\ref{fig:new_gated_conv_mask}}.

Although face images can be desensitized by adding the mask matrix, they may still be restored by face inpainting methods. Face inpainting methods aim to restore damaged images to a reasonably visual one. To observe the influence of inpainting images for recognition, all masked face images are restored by RFR-Net \cite{li2020recurrent}, a SOTA image inpainting method. The result is shown in {Fig.~\ref{fig:new_gated_conv_mask}} and {Table~\ref{tab:searched_mask_nvidia}}. As shown in {Table~\ref{tab:searched_mask_nvidia}}, after face inpainting, the recognition accuracy of resnet significantly improves while the resnet-mask slightly declines. This phenomenon indicates that in the forward propagation of the backbone network, unknown areas are filled with features generated by the convolution layer, which are different from those in the inpainting model. This difference may result in the recognition accuracy decline in resnet-mask after image inpainting. Even though the results from {Table~\ref{tab:searched_mask_nvidia}} show that the recognition accuracy of resnet-mask declines after inpainting, the RFR-Net can still restore the missing pixels well in the view of the human being (as shown in Fig.~\ref{fig:new_gated_conv_mask}). This may be incurred by the low mask ratio of the masks.

\begin{table}[t]
\centering
\caption{The recognition accuracy of resnet and resnet-mask, the input image is desensitized by the mask, which searched in the mask level (0.4-0.5) and (0.5-0.6).}
\setlength{\tabcolsep}{1mm}{
    \begin{tabular}{@{}lcccccc@{}}
    \toprule
    Mask level & 
    \multicolumn{2}{@{}c@{}}{resnet} & \multicolumn{2}{@{}c@{}}{resnet-mask} & Ratio\\
    \midrule
    & masked & inpainting & masked & inpainting & \\
    \midrule
0.4-0.5 & 78.02\% & 89.9\% & 90.23\% & 88.58\% & 42.35\%\\

0.5-0.6 & 63.57\% & 78.82\% & 85.82\%  & 81.57\% & 50.94\%\\
    \bottomrule
    \end{tabular}}
    \label{tab:searched_mask_nvidia}
\end{table}

\subsection{Results of the proposed method}
\label{subsec:result_searched_mask}
We design a series of experiments in this subsection to validate our proposed method. 

\begin{figure*}[t]
  \centering
  \includegraphics[width=1\linewidth]{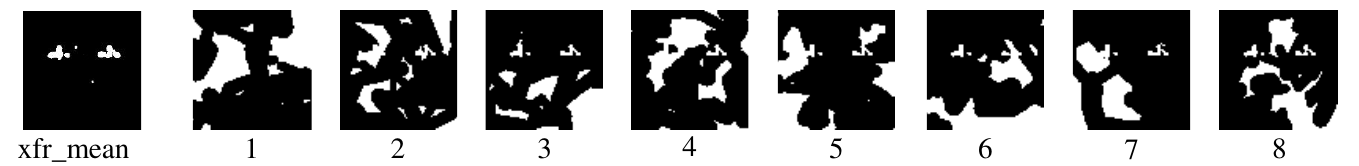}
  \caption{Examples of searched masks and mean saliency map with {Algorithm.~\ref{algos:mask_search}}. The left is the mean saliency map under thershold=0.5, and the numbers 1-8 are searched masks. The best mask is number 8.}
  \label{fig:searched_mask}
\end{figure*}

\textbf{Mask search and recognition with FSM.} We exploit Algorithm. \ref{algos:mask_search} to search the mask with the hyper-parameter $n=8$, and EBP threshold $T=0.5$. All searched masks are shown in {Fig.~\ref{fig:searched_mask}}. The mask with the lowest value of Formula \ref{eq:mask_optimize} is shown in the bottom right corner of {Fig.~\ref{fig:searched_mask}}. We combine FSM and the backbone network after Layer 3 in resnet. Face recognition accuracy with pixel masking is shown in {Table~\ref{tab:searched_mask}}. As shown in {Table~\ref{tab:searched_mask}}, the proposed FSM can improve face recognition accuracy (up to 9.34\% compared to the baseline model) by gating filled pixels in masked areas.

\begin{table*}[t]
\centering
\caption{The face recognition accuracy with different models (the best searched mask is shown in {Fig.~\ref{fig:searched_mask}}).}
\setlength{\tabcolsep}{1.5mm}{
    \begin{tabular}{@{}cccc@{}}
    \toprule
    {resnet} & {resnet-mask} & {resnet-fsm} & Mask ratio\\
    \midrule
    53.92\% & 79.78\% & \textbf{89.12\%} & 76.03\%\\
    \bottomrule
    \end{tabular}}
    \label{tab:searched_mask}
\end{table*}

\begin{figure*}[t]
  \centering
  \includegraphics[width=0.8\linewidth]{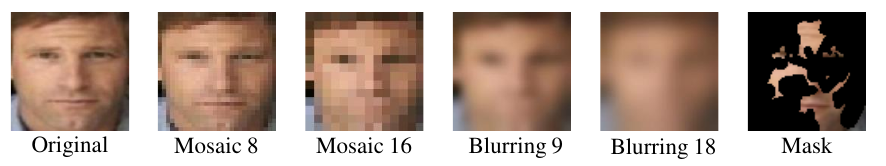}
  \caption{Examples of different desensitization algorithms. We show the face images using mosaic with kernel sizes 8, and 16, the face images using Gaussian blurring with kernel sizes 9 and 18, and the face image with the searched mask.}
  \label{fig:blurring_mosaic}
  \vspace{-6mm}
\end{figure*}

\begin{sidewaystable}
\sidewaystablefn%
\begin{center}
\begin{minipage}{\textheight}
\caption{The recognition accuracy of resnet, resnet-trained, and res net-fsm. $^{*}\rm Higher$ is better.}
\setlength{\tabcolsep}{1mm}{
    \begin{tabular}{@{}lcccccccccccc@{}}
    \toprule
    Model & 
    \multicolumn{2}{c}{Blurring \cite{raynal2020image}} & \multicolumn{2}{c}{Mosaic \cite{raynal2020image}}
    & \multicolumn{6}{c}{Nvidia Mask \cite{liu2018partialinpainting}}& \multicolumn{1}{c}{Ours}\\
    \midrule 
    & \multicolumn{2}{c}{Kernel Size} & \multicolumn{2}{c}{Kernel Size}  & \multicolumn{6}{c}{Mask Levels} & \multicolumn{1}{c}{}\\
    & 9 & 18 & 8 & 16 & 1 & 2 & 3 & 4 & 5 & 6 & \\
    \midrule
    resnet & 79.13\% & 70.13\% & 85.65\% & 73.1\% & 90.3\% & 83.71\% & 73.54\% & 62.08\% & 56.03\% & 53.63\% & 53.92\%\\

    resnet-trained & 93.92\% & 65.88\% & 94.45\% & 86.93\% & 91.89\% & 91.32\% & 90.49\% & 89.53\% & 87.76\% & 83.62\% & 89.12\%\\
    
    $\rm dSSIM^{*}$ & 0.21 & 0.29 & 0.34 & 0.42 & - & - & - & - & - & - & 0.81\\
    \bottomrule
    \end{tabular}}
    \label{tab:different_desentations}
\end{minipage}
\end{center}
\end{sidewaystable}


    

\textbf{Compare with diverse desensitization methods.} Besides masks, some other visual image desensitization methods exist, such as blurring and mosaic \cite{raynal2020image}. Blurring is widely exploited to remove noise and reduce high-frequency information in images. The most used blurring technique is Gaussian blurring \cite{raynal2020image}, which utilizes a sliding Gaussian kernel to convolve the entire image. Mosaic is usually exploited to reduce details in images, reducing resolution while keeping the image size as the original one. With mosaic technology, the image is first divided into a square grid, then pixels in each grid are set to one value, e.g., the average of the pixels in that grid. Examples of different desensitization technologies are shown in {Fig.~\ref{fig:blurring_mosaic}}. While mosaic and blurring destroy many details in face images, the contours of images are still preserved, and some sensitive information may still exist.

Similar to {Section~\ref{subsec:exp_settings}}, we denote the model trained on clean face images as resnet. With this notation, we exploit resnet-trained to represent resnet-blurring, resnet-mosaic, and resnet-mask for convenience. We utilize dSSIM \cite{wang2004image} to compare different desensitization methods. SSIM reflects the spatial correlation between the desensitized images and the original images, and $dSSIM=1-SSIM$ could reflect the privacy of the desensitized image \cite{raynal2020image}. A high value of the dSSIM corresponds to better privacy protection. The results of different methods are shown in {Table~\ref{tab:different_desentations}}. This experiment shows that the proposed mask-based desensitization method corresponds to the best privacy result.

\textbf{Different test situations.} According to the input image pair, there are three test situations on the face recognition with pixel masking model.
\begin{itemize}
    \item Situation 1: The query image and the image in the gallery are both masked.
    \item Situation 2: The query image is not masked, and the image in the gallery is masked.
    \item Situation 3: Both the query image and the image in the gallery are not masked.
\end{itemize}

\begin{figure}[t]
  \centering
  \includegraphics[width=0.6\linewidth]{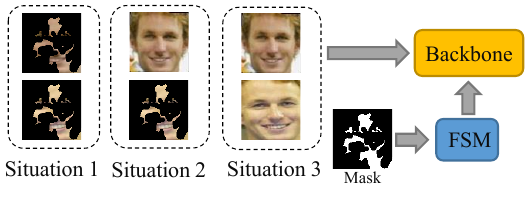}
  \caption{Different test situation.}
  \label{fig:different_test_situations}
\end{figure}

The illustration of these three situations are shown in {Fig.~\ref{fig:different_test_situations}}. The recognition accuracy with three situations is shown in {Table~\ref{tab:three_situation}}. Although the model is trained on the masked face, it maintains an excellent recognition accuracy of clean face images (situation 3).

\rev{According to Table~\ref{tab:different_desentations}, we find that compared to existing image desensitization techniques such as blurring and mosaic, our method improves the dSSIM value by 0.52 and 0.39, respectively, which proves using a large-ratio mask to occlude the image can more effectively protect the privacy information of faces. Simultaneously, by employing our proposed FSM network, we can address the issue of data unavailability caused by the introduction of mask. According to Table~\ref{tab:searched_mask}, the method incorporating the FSM network (resnet-fsm) achieves a 9.34\% improvement in accuracy compared to the method without the FSM network (resnet-mask) when recognizing images with mask.}

\begin{table}[t]
\centering
\caption{The resnet-fsm recognition accuracy on different situations.}
    \begin{tabular}{@{}cccc@{}}
    \toprule
    Model & {Situation 1} & {Situation 2} & {Situation 3}\\
    \midrule
    resnet-fsm & 89.12\% & 89.37\% & 91.35\% \\
    \bottomrule
    \end{tabular}
    \label{tab:three_situation}
\end{table}

\subsection{Ablation study}
\label{subsec:ablation}
\begin{figure}[t]
  \centering
  \includegraphics[width=0.5\linewidth]{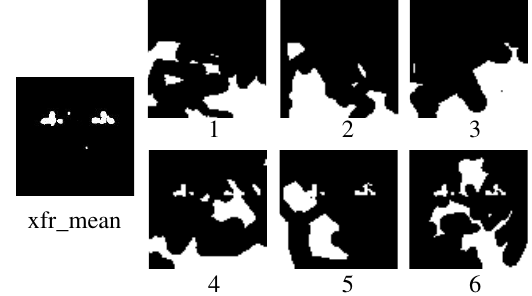}
  \caption{Examples of some searched masks with and without EBP. Numbers 1-3 are searched masks without EBP; numbers 4-6 are searched masks with EBP.}
  \label{fig:with_and_without_ebp}
\end{figure}

\textbf{Desensitization without EBP.} As mentioned in {Section~\ref{subsec:mask_search}}, to reduce the time consumption of the search procedure and improve search efficiency, we exploit EBP to guide the mask generation and search. To prove the importance of EBP in the desensitization procedure, we show some searched masks with and without EBP in {Fig.~\ref{fig:with_and_without_ebp}}. The recognition accuracy of some searched masks and PSNR are shown in {Table~\ref{tab:with_and_without_ebp}}. PSNR is a quantization metric for measuring the restoration result after inpainting. With the guarantee of EBP, searched masks (numbers 4-6) are better than those without EBP (numbers 1-3) in the recognition accuracy of resnet-mask. Interestingly, the inpainting result improves when searching for masks with EBP. This may indicate that an excellent inpainting algorithm relies on essential areas that contribute more to the recognition algorithm, e.g., the eye areas.

\begin{table}[t]
\centering
\caption{The face recognition accuracy with different masks. $^{*}\rm Higher$ is better.}
\setlength{\tabcolsep}{0.6mm}{
    \begin{tabular}{@{}ccccccc@{}}
    \toprule
    Model & 1 & 2 & 3 & 4 & 5 & 6\\
    \midrule
    resnet & 56.1\% & 54.3\% & 54.2\% & 56.1\% & 55\% & 53.9\%\\
    resnet-mask & 72.8\% & 71.5\% & 75.2\% & 76.8\% & 74.9\% &79.8\%\\
    $\rm PSNR^{*}$ & 15.8 & 14 & 13.7 & 19.3 & 18.6 & 18.4 \\
    Mask ratio & 79.4\% & 80.5\% & 73.6\% & 74.7\% & 79.5\% & 76\%\\
    \bottomrule
    \end{tabular}}
    \label{tab:with_and_without_ebp}
\end{table}

\textbf{Face recognition without FSM.} In our experiment, resnet-mask refers to the results without the FSM. The results in Table~\ref{tab:searched_mask_nvidia} and Table~\ref{tab:searched_mask} show that without the FSM, the recognition accuracy under mask can be inferior compared with resnet-fsm.

\textbf{FSM insertion point.} Because FSM consists of stacked basic modules, the size of the output feature map can be adjusted according to the number of stacked modules so that it can be inserted anywhere in the backbone network. The resnet network has four modules: Layer 1, Layer 2, Layer 3, and Layer 4. The corresponding output feature map size is 56, 28, 14, and 7. Therefore, we inserted FSM after these four layers respectively to explore the feature-selected capability of FSM under different insertion points. The recognition accuracy of different insert point is shown in Table~\ref{tab:fsm_insert_point}.

\begin{table}[t]
\centering
\caption{The resnet-fsm recognition accuracy on different situations.}
    \begin{tabular}{@{}ccccc@{}}
    \toprule
    Model & Layer1 & Layer2 & Layer3 & Layer4\\
    \midrule
    resnet-fsm & 87.36\% & 88.73\% & \textbf{89.12\%} & 88.92\% \\
    \bottomrule
    \end{tabular}
    \label{tab:fsm_insert_point}
\end{table}

\begin{figure*}[t]
  \centering
  \includegraphics[width=0.65\linewidth]{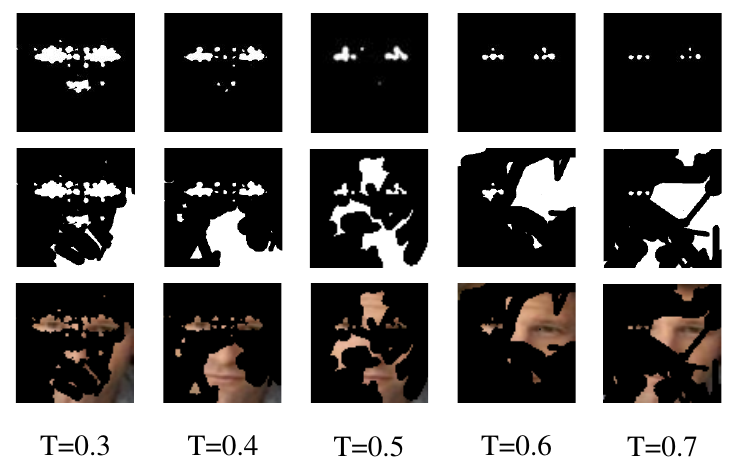}
  \caption{\rev{The xfr\_mean obtained at different T values and the corresponding mask found at each T value. The first row represents the xfr\_mean obtained at different T values, while the second row represents the optimal mask selected according to the Algorithm.~\ref{algos:mask_search} for each corresponding xfr\_mean.}}
  \label{fig:differentT}
\end{figure*}

\rev{\textbf{Desensitization with different T.} The threshold value T directly affects the visible area in the xfr\_mean. A larger T results in a smaller visible area in the xfr\_mean, whereas a smaller T leads to a larger visible area. The size of the visible area in the xfr\_mean directly impacts the difficulty of searching for masks and the privacy of the found masks. In order to investigate the impact of different values of T on the desensitization algorithm, we conduct experiments to search for the optimal mask under different binarization threshold values T, as shown in the Fig~\ref{fig:differentT} and Table~\ref{tab:differentT}.  We find that setting T to 0.5 achieves the optimal balance between privacy and recognition accuracy.}

\begin{table*}[t]
\centering
\caption{\rev{The dSSIM of the mask found at different T values and its recognition accuracy in the resnet-fsm model.}}
\setlength{\tabcolsep}{1.5mm}{
    \begin{tabular}{@{}cccc@{}}
    \toprule
    {T} & {resnet-mask} & {resnet-fsm} & dSSIM\\
    \midrule
    0.3 & 79.4\% & 86.42\% & 0.76\\
    \midrule
    0.4 & 79.8\% & 88.6\% & 0.77\\
    \midrule
    0.5 & 79.78\% & \textbf{89.12\%} & \textbf{0.81}\\
    \midrule
    0.6 & 76.05\% & 83.68\% & 0.78\\
    \midrule
    0.7 & 77\% & 84.8\% & 0.78\\
    \bottomrule
    \end{tabular}}
    \label{tab:differentT}
\end{table*}

\section{Conclusion}
\label{sec:conclusion}
In this work, we propose a mask-based image desensitization approach, which consists of an image desensitization algorithm and FSM. The image desensitization algorithm can generate proper masks to destroy images' visual information while preserving important images' important features for recognition. The feature select masknet can improve the accuracy of recognizing desensitized face images. Furthermore, we extend these two algorithms to federated learning scenarios, where the desensitized face images are only exploited locally to further protect the data's privacy. Extensive experimental results demonstrate that our approach can significantly outperform (up to 9.34\%) the state-of-the-art baseline approaches in terms of recognition accuracy and the best privacy compared to the mosaic-based and blurring-based desensitization methods. 

\rev{While our approach is effective to protect the sensitive information of images, it cannot generate personalized mask for each input image. As a result, it can only be used for images under constrained conditions, where the face features in each image are relatively fixed in position. Our future work focus on designing a desensitization approach for arbitrary images.}

\section*{Acknowledgements}

We acknowledge the discussion with Dr. Yulun Zhang, who provided valuable comments while preparing the manuscript.

\section*{Declarations}

Data available on request from the authors.

\noindent The authors declare that there is no conflict of interest.




\bibliography{bib}


\end{document}